\PassOptionsToPackage{table}{xcolor}
\documentclass[11pt]{tibop-article}
\usepackage[utf8]{inputenc}
\usepackage[T1]{fontenc}
\usepackage{geometry}
\usepackage{hyperref}
\usepackage{subcaption} 
\usepackage{enumitem}  
\usepackage{listings}  
\usepackage{booktabs} 
\definecolor{alexcolor}{HTML}{FFDD88} 
\usepackage{siunitx}  
\usepackage{multirow} 
\usepackage{amsmath}

\lstdefinelanguage{json}{
 basicstyle=\ttfamily\small,
 numbers=left,
 numberstyle=\scriptsize,
 stepnumber=1,
 frame=single,
 showstringspaces=false,
 breaklines=true,
 literate=
 *{0}{{{\color{black}0}}}{1}
 {1}{{{\color{black}1}}}{1}
 {2}{{{\color{black}2}}}{1}
 {3}{{{\color{black}3}}}{1}
 {4}{{{\color{black}4}}}{1}
 {5}{{{\color{black}5}}}{1}
 {6}{{{\color{black}6}}}{1}
 {7}{{{\color{black}7}}}{1}
 {8}{{{\color{black}8}}}{1}
 {9}{{{\color{black}9}}}{1}
 {:}{{{\color{black}{:}}}}{1}
 {,}{{{\color{black}{,}}}}{1}
 {\{}{{{\color{black}{\{}}}}{1}
 {\}}{{{\color{black}{\}}}}}{1}
 {[}{{{\color{black}{[}}}}{1}
 {]}{{{\color{black}{]}}}}{1},
}
\TIBauthor[A. Beliaeva and T. Rahmatullaev]{Aleksandra Beliaeva *\affOne\affTwo\affThree\orcidlink{0009-0005-4627-3666}\and
   Temurbek Rahmatullaev *\affFour\affFive\orcidlink{0000-0001-5570-0010}
   }
\TIBaffiliations{
\affOne Skolkovo Institute of Science and Technology, Russia\medskip\\
	\affTwo BIMAI-Lab, Biomedically Informed Artificial Intelligence Laboratory, University of Sharjah, Sharjah, United Arab Emirates\medskip\\
        \affThree Ivannikov Institute for System Programming of the Russian Academy of Sciences, Russia\medskip\\
        \affFour Lomonosov Moscow State University, Russia\medskip\\
        \affFive AIRI Institute, Russia\medskip\\
	*These authors contributed equally and are joint first authors.\medskip\\Correspondence: belyaeva.alex1@gmail.com, raxtemur@gmail.com
	
}

\TIBtitle{Alexbek at LLMs4OL 2025 Tasks A, B, and C:}
\TIBsubtitle{Heterogeneous LLM Methods for Ontology Learning\\(Few-Shot Prompting, Ensemble Typing, and Attention-Based Taxonomies)}
\TIBbundlename[ConfAbbrev]{Full conference name}
\TIBbundlename[Open Conf Proc X (2025) "LLMs4OL 2025: The 2nd Large Language Models for Ontology Learning Challenge at the 24th ISWC"]{LLMs4OL 2025: The 2nd Large Language Models for Ontology Learning Challenge at the 24th ISWC}
\TIBconferencesession{LLMs4OL 2025 Task Participant Papers}
\TIBabstract{We present a comprehensive system for addressing Tasks A, B, and C of the LLMs4OL 2025 challenge, which together span the full ontology construction pipeline: term extraction, typing, and taxonomy discovery. Our approach combines retrieval-augmented prompting, zero-shot classification, and attention-based graph modeling — each tailored to the demands of the respective task. 

For Task A, we jointly extract domain-specific terms and their ontological types using a retrieval-augmented generation (RAG) pipeline. 
Training data was reformulated into a document to terms and types correspondence, while test-time inference leverages semantically similar training examples. This single-pass method requires no model finetuning and improves overall performance through lexical augmentation

Task B, which involves assigning types to given terms, is handled via a dual strategy. In the few-shot setting (for domains with labeled training data), we reuse the RAG scheme with few-shot prompting. In the zero-shot setting (for previously unseen domains), we use a zero-shot classifier that combines cosine similarity scores from multiple embedding models using confidence-based weighting.

In Task C, we model taxonomy discovery as graph inference. Using embeddings of type labels, we train a lightweight cross-attention layer to predict is-a relations by approximating a soft adjacency matrix.

These modular, task-specific solutions enabled us to achieve top-ranking results in the official leaderboard across all three tasks. Taken together these strategies showcase the scalability, adaptability, and robustness of LLM-based architectures for ontology learning across heterogeneous domains. 

Code is available at: \url{https://github.com/BelyaevaAlex/LLMs4OL-Challenge-Alexbek}
}
\TIBkeywords{Large Language Models, Ontology engineering (OE), Ontology Learning, Domain-specific knowledge, Retrieval Augmented Generation, Term Typing, Taxonomy Discovery}
\usepackage{xfrac}
\begin{document}
\maketitle

\section{Introduction}

Large Language Models (LLMs) have recently emerged as powerful tools for ontology learning (OL) tasks, including term extraction, typing, and taxonomy discovery. Prior studies have demonstrated that, when combined with retrieval-augmented generation (RAG) and in-context learning techniques, LLMs can act as general-purpose engines for structured knowledge extraction, even in the absence of task-specific finetuning \cite{giglou2023llms4ollargelanguagemodels,phoenixes}. Three complementary strategies have proven especially effective in this context: (i) few-shot prompting with curated examples, (ii) embedding-based similarity search, and (iii) zero-shot classification using pretrained models.

A notable illustration of this paradigm is the LLMs4OL framework \cite{giglou2023llms4ollargelanguagemodels}, which applies zero-shot prompting across three core OL subtasks—term typing, taxonomy induction, and non-taxonomic relation extraction—evaluated over diverse domains. The results suggest that, while base LLMs exhibit strong baseline capabilities, task-specific finetuning significantly improves performance. Building on this foundation, the LLMs4OL 2024 Challenge introduced a standardized benchmark encompassing few-shot and retrieval-augmented settings across five domains and a set of structured tasks (A/B/C) \cite{giglou2024llms4ol2024overview1st,phoenixes}.

Extending these ideas further, OLLM \cite{lo2024endtoendontologylearninglarge} reframes OL as an end-to-end fine-tuned process, integrating term and taxonomy learning into a unified architecture. Through the use of a custom regularizer and novel graph-aware evaluation metrics, OLLM achieves state-of-the-art results on both general (Wikipedia) and specialized (arXiv) corpora. Together, these developments reflect a shift in the field—from isolated prompting strategies toward integrated, data-grounded pipelines that scale across domains.

Complementing this trend, domain-specific approaches like LLMs4Life \cite{fathallah2024llms4lifelargelanguagemodels} 
show how careful prompt engineering and ontology reuse yield richer hierarchies and more consistent ontologies in specialized domains.

Despite these advances, prior systems often rely on heavy finetuning or manually designed prompts, and lack a unified, efficient pipeline that works across tasks and domains.
Our aim is to design a modular, lightweight, and scalable system that tackles all three OL tasks of the LLMs4OL 2025 benchmark:
\begin{itemize}
\item \textbf{Task A: Text2Onto} — joint extraction of terms (A1) and their types (A2) from raw domain documents.
\item \textbf{Task B: Term Typing} — assigning types to given terms, evaluated in both \emph{few-shot} (B1–B3) and \emph{zero-shot} (B4–B6) settings.
\item \textbf{Task C: Taxonomy Discovery} — inducing \emph{is-a} relationships between types to construct hierarchical taxonomies (C1–C8).
\end{itemize}
We target a single unified framework that avoids full encoder finetuning, yet remains competitive.

\vspace{0.5em}
\noindent
\textbf{Our contributions are threefold:}
\begin{enumerate}[noitemsep,topsep=2pt]
  \item We present a unified, modular LLM-based system for ontology learning, covering term extraction, term typing, and taxonomy induction across Tasks A, B, and C of the LLMs4OL 2025 challenge.
  \item We introduce a simple yet effective \textbf{dedicated cross-attention layer} for taxonomy discovery, which enables efficient graph inference over type embeddings without full encoder finetuning.
  \item Our method achieves competitive results across all subtasks, including top-2 placement in multiple domains and strong zero-shot generalization. It is fully task-agnostic, lightweight, and requires no large-scale finetuning.
\end{enumerate}

This work challenges the prevailing notion that OL requires heavy finetuning or complex multi-component pipelines. We show that a lean, modular system—built on efficient prompting and adaptive lightweight modules—can compete with or outperform more expensive methods, supporting scalable, domain-agnostic ontology learning in real-world settings.

\section{Methodology}

\subsection{Task A: Text2Onto}
\label{sec:method:taskA}

Task~A entails concurrent extraction of \emph{terms}
(SubTask~A1) and their \emph{types} (SubTask~A2) from raw documents
spanning the \textbf{Ecology}, \textbf{Scholarly}, and
\textbf{Engineering} domains.
We resolve both subtasks in a single pass by
(i)~curating the released data and
(ii)~employing retrieval-augmented few-shot prompting.

\begin{figure}[ht]
  \centering
  \includegraphics[width=0.9\textwidth]{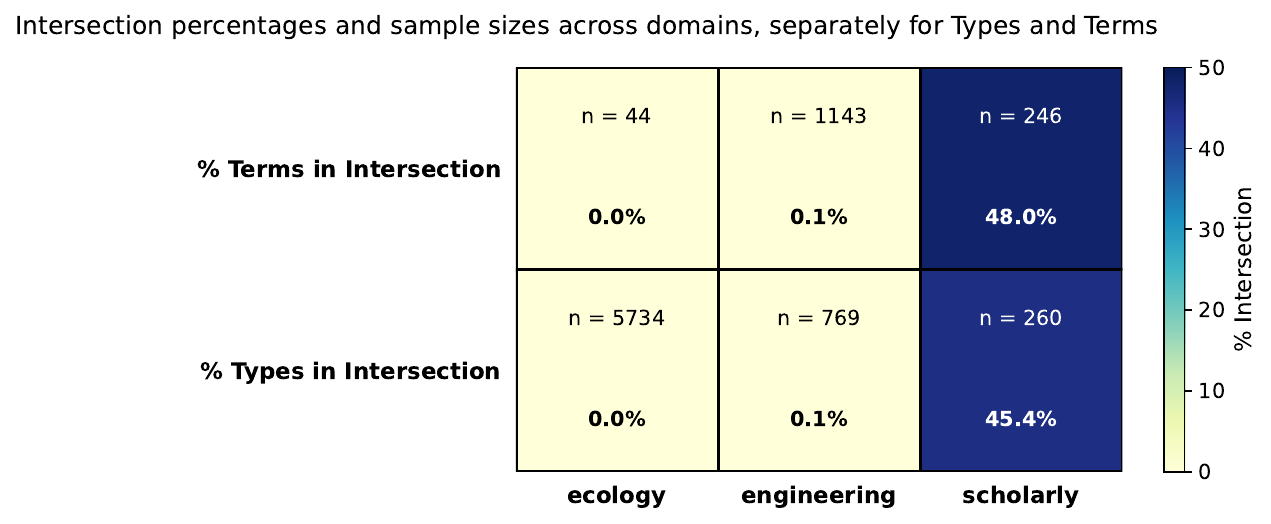}
  \caption{
    Normalized term–type intersection across domains.
    Overlap is near-zero in \textbf{Ecology} and \textbf{Engineering}, 
    indicating weak supervision, while \textbf{Scholarly} shows strong alignment, 
    with nearly half of terms and types intersecting.
  }
  \label{fig:term-intersection}
\end{figure}

\paragraph{Dataset inspection and repair}
The official distribution comprises the files:
\texttt{documents.jsonl}, \texttt{terms.txt}, \texttt{types.txt}, 
\texttt{terms2docs.json}, and \texttt{terms2types.json}. A manual audit revealed that
\texttt{terms2docs.json} actually maps \emph{types} to document
IDs. To assess the issue, we measured term–type overlap across domains (see Fig.~\ref{fig:term-intersection}). %

To correct the issue, we rescanned documents for exact matches from \texttt{terms.txt}, reconstructing a term–document index.  
Merging it with \texttt{terms2types.json} yields reliable supervision tuples  
$\langle\!\textit{doc},\,\{\textit{term},\textit{type}\}\!\rangle$.

\paragraph{Instruction reformulation}
Each training document is converted to an
\texttt{\small <INSTRUCTION>}\,/\,\texttt{\small <OUTPUT>} pair.
The input concatenates
\textbf{(a)}~title,
\textbf{(b)}~full text, and
\textbf{(c)}~the top-20 TF–IDF keywords;
the output is
\verb|{"terms":[...], "types":[...]}|.
Early ablation showed that TF–IDF augmentation increases recall by $\approx 1.3$\,pp, so we retain it in the final pipeline.

\paragraph{Embedding-based retrieval}
For every test document, we retrieve the
$k\!=\!3$ nearest neighbours (cosine similarity) from the training
corpus—a value chosen to balance diversity and prompt length.
All documents are embedded once using
\textbf{Qwen3-Embedding-4B}\cite{qwen3embedding}, which supports 32k-token input and
offers strong multilingual semantics. While larger variants (e.g., 8B)
may improve quality, the 4B model provides a good trade-off between
performance and cost. Retrieved examples are prepended as few-shot
demos to guide the LLM in generating terms and types.

\paragraph{Prompt configurations} 

For each dataset in SubTask~1 we evaluate two few‐shot prompting strategies, hereafter referred to as \textbf{Method 1} and \textbf{Method 2}:
\begin{enumerate}
  \item \textbf{Method 1 (Doc$\rightarrow$Type exemplars).} Few‐shot examples are built directly from \texttt{terms2docs.json}. Although the file name suggests a term–document index, it in fact maps \emph{types} to document IDs; every demonstration therefore corresponds to a \emph{doc2type} pair.
  \item \textbf{Method 2 (Chained Term\&Type exemplars).} Few‐shot examples combine both terms and types, obtained by chaining \texttt{terms2types.json} with the reconstructed \texttt{terms2docs.json}.
\end{enumerate}

For the same datasets we then perform type extraction under two configurations:
\begin{enumerate}
  \item \textbf{Method 1 (Term$\rightarrow$Type continuation).} We take the terms extracted in the previous subtask and augment them with few‐shot examples built from \texttt{terms2types.json}.
  \item \textbf{Method 2 (Chained Term\&Type exemplars).} Identical to Method 2 above.
\end{enumerate}

\paragraph{Generation and post-processing}
Structured output is generated using the Outline library\cite{yoshino2024outline}, which enforces a JSON-like format during decoding.
The resulting predictions are concatenated, deduplicated, and saved to  
\texttt{terms.txt} (A1) and \texttt{types.txt} (A2).  
The entire pipeline runs in a single pass and requires no parameter tuning.

\subsection{Task B: Term Typing}
\textit{Assigning an ontological type to each lexical term}

Task~B involves predicting ontology types for individual lexical terms, split across two tracks:  
(i)~few-shot learning on training sets from known domains (SubTasks~B1–B3: \textbf{OBI} (biomedicine), \textbf{MatOnto} (materials science), and \textbf{SWEET} (earth sciences)), and  
(ii)~zero-shot generalisation to unseen domains (SubTasks~B4–B6).

\subsubsection{Few‑shot (subtasks B1–B3)} 

\paragraph{Retrieval-augmented prompting}

We follow the same few-shot strategy as in Task~A.  
All training terms are embedded using \textbf{Qwen3-Embedding-4B}, and for each test term, we retrieve its top-$k$ ($k\!=\!3$) nearest neighbors by cosine similarity.  
The retrieved \{\textit{term}, \textit{types}\} pairs are inserted into the prompt as demonstrations.  
This helps the LLM assign ontological types by analogy.  
Ablation confirmed that retrieval improves recall by $\approx1.3$\,pp, and we retain it in the final pipeline.

\paragraph{Embedding-based baselines}

Additionally, to assess the value of few-shot prompting explicitly, we compared it against embedding-based baselines: (1) a Random Forest classifier trained on pretrained Qwen3 embeddings, and (2) the same classifier augmented with graph-derived features from a term co-occurrence graph.

\subsubsection{Zero‑shot (subtasks B4–B6)} 
\paragraph{Cosine similarity}
Each test term and all candidate types are embedded using either  
\textbf{MPNet} (\texttt{all-mpnet-base-v2}, 768-dim, mean pooling)  
or \textbf{Qwen3-Embedding-4B} (2560-dim, last-token pooling).  
We assign to each term the type with highest cosine similarity.  
Embeddings are preprocessed via L2 normalisation (batch size 32 for Qwen).  
Prompt styles tested include: \textit{plain} (no context), \textit{QA-style}, and \textit{instructional} — applied consistently to both terms and types. MPNet was selected for its strong STS performance; Qwen3 provides long-context generalisation.   Together, they offer complementary views of embedding space.

\paragraph{Ensemble classifier}
To combine model strengths, we use a dynamically weighted ensemble of  
MPNet, Qwen3, and \textbf{BGE} (\texttt{bge-large-en-v1.5}).  
Weights are computed per sample based on prediction confidence:
\begin{enumerate}[label=\arabic*. ,itemsep=4pt,topsep=2pt]
  \item \textbf{Confidence score.}  
        For each model, compute  
        \(\textit{confidence}=p_{\max}(1 - H_{\text{norm}})\),  
        where \(p_{\max}\) is the max softmax probability, and \(H_{\text{norm}}\) is the normalised entropy.
  \item \textbf{Weight update.}  
        Final weights are assigned as  
        \(\textit{weight} = 0.7\,\textit{confidence} + 0.3(1 - H_{\text{norm}})\),  
        then normalised so that \(\sum w = 1\).  
        This heuristic yielded more stable predictions than confidence-only weighting.
  \item \textbf{Domain-specific prompting.}  
        Retrieved examples are prepended using domain-adapted templates:
        \begin{itemize}[noitemsep]
          \item B4: \texttt{"In Earth sciences, explain..."}
          \item B5: \texttt{"In linguistics, define..."}
          \item B6: \texttt{"In metrology, describe..."}
        \end{itemize}
        Type prompts follow: \texttt{"This <domain> category represents/encompasses"}.
  \item \textbf{Aggregation.}  
        Final prediction is made by summing similarity scores across models, weighted by their dynamic weights.
\end{enumerate}

\paragraph{DistMult-based scoring}
We additionally evaluate a compositional method based on \textbf{DistMult}, where each term--type pair \((t, y)\) is scored via dot product of their mean-pooled Qwen3-Embedding-4B representations:
\[
\text{score}(t, y) = \mathbf{t}^\top \mathbf{y}
\]
Predicted types are selected using adaptive z-score thresholding.  
Unlike cosine similarity, this formulation supports multi-type predictions and captures latent compatibility beyond lexical overlap.

\subsection{Task~C: Taxonomy Discovery\texorpdfstring{\vspace{2pt}}{}}
\textit{Identifying hierarchical (\textit{is‐a}) relations between types}

\paragraph{Task C Overview}
Task~C consists of eight domains (C1–C8): OBI, MatOnto, SWEET, DOID, Schema.org, PROCO, FoodOn, and PO.  
Each domain provides a flat list of types and a ground-truth taxonomy  
 of parent–child (\textit{is-a}) relations. At test time, only the type list is available; participants must infer the missing hierarchy.

\paragraph{Problem setup}
We model taxonomy discovery as prediction of an \emph{adjacency
matrix} \(A \in \mathrm{Mat}(N,N)\) where \(N\) is the number of
types and \(a_{ij}\!\in\![0,1]\) denotes the likelihood that
\(\textit{type}_i\) is a subclass of \(\textit{type}_j\).
The goal is to infer a directed graph whose edges align with the
ground-truth \textit{is-a} relations.

\paragraph{Model}

\begin{figure}[ht]                    
  \centering
  \includegraphics[width=1\textwidth]{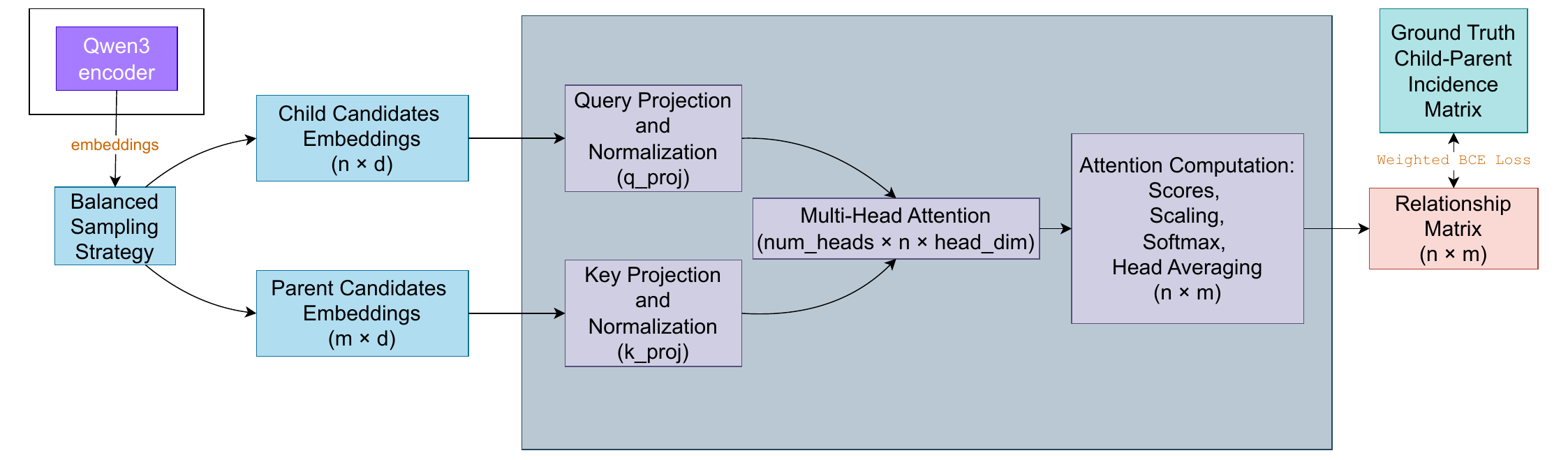}
  \caption{Cross-Attention Architecture for Taxonomy Discovery. Child and parent candidate embeddings (from Qwen3) are projected into query and key spaces, respectively; a multi-head attention block computes pairwise scores, which after scaling and softmax yield a predicted child-parent relation matrix. This matrix is trained against the ground-truth incidence graph using a weighted binary cross-entropy loss.}
  \label{fig:tasC_method}      
\end{figure}

We implement a single \textbf{cross-attention layer} that maps two
identical term sequences \(\mathrm{TERMS}_1\) and
\(\mathrm{TERMS}_2\) to an attention matrix
\(\hat A \in [0,1]^{N\times N}\), as illustrated in Fig.~\ref{fig:tasC_method}. This approach is simple but showed its effectiveness.

The parameters are optimised with binary cross-entropy against
the incidence matrix of the true taxonomy. 

We study two encoder–attention configurations. 
\textbf{(i) 4B‑Frozen.} Qwen3‑Embedding‑4B is kept frozen; a single
cross‑attention layer is trained on top of its type embeddings.  
\textbf{(ii) 0.6B+LoRA.} A lighter Qwen‑0.6B encoder is equipped with
rank‑8 LoRA adapters, ($\alpha$ = 16) and \emph{jointly} updated
together with the same cross‑attention head.  In both cases the encoder
is applied twice to the identical sequence of type tokens
$\mathrm{TERMS}_1=\mathrm{TERMS}_2$, and the resulting contextual
representations are projected to the attention matrix
$\hat A \in [0,1]^{N \times N}$. 

\paragraph{Training strategy comparison}
To evaluate the trade‐off between embedding size and adaptability, we compare two training configurations:
\begin{itemize}[noitemsep,topsep=2pt]
  \item \textbf{Frozen 4B + Cross‐Attn:} the Qwen-4B embedder remains frozen and only the cross‐attention layer is trained.
  \item \textbf{LoRA‐Finetuned 0.6B + Cross‐Attn:} we apply LoRA to a smaller Qwen-0.6B embedder and jointly train it with the cross‐attention layer.
\end{itemize}

\paragraph{Training}
Each domain is split 80 / 20 into train and validation sets by types rather than edges; only \textit{is-a} pairs with both terms in the same partition are retained. For 4B‑Frozen we grid‑search the
learning rate ($\{1,5\}\!\times\!10^{-5}$), batch size
$\{4,8,16,32\}$, number of heads $\{4,8,32\}$ and epochs (2–15).  
For 0.6B+LoRA we only search among epochs amount $\{2,5,15,50,100\}$ and a
fixed starting LR $5\times10^{-6}$. We employ a cosine learning‐rate schedule with a linear warm‐up phase covering 10 \% of the total training steps.
The binary‑cross‑entropy loss is averaged over all valid $(i,j)$ pairs,
counting missing edges as negatives.

\paragraph{Evaluation}
Checkpoints are first ranked by validation ROC AUC.  On the test set we
compaired two thresholding schemes:  
(a) \emph{Validation‑F$_1$} — threshold that maximises F$_1$ on the
validation split;  
(b) \emph{Sparsity‑Matched} — threshold that replicates the edge
density observed in the training graph (i.e.\ the ratio
\(\lvert\!\text{pred.\ edges}\rvert / N^{2}\)).  The latter is used as the
default in the main leaderboard, while Validation‑F$_1$ for 0.6B+LoRA is included for ablation.

\section{Results}

See final competition results in Appendix A.

\subsection{Task A: Terms and Types Extraction}

\subsubsection{Terms Extraction}

Table \ref{tab:subtaskA1-results} contrasts both methods on the \textbf{Scholarly} and \textbf{Engineering} datasets. Best scores per column are highlighted in bold: Method 2 dominates on all metrics in the Scholarly domain, whereas Method 1 retains an edge in $F_{1}$ and precision for Engineering, despite minimal term–type overlap in this domain (cf. Fig.~\ref{fig:term-intersection}).

\begin{table}[ht]
\centering
\caption{SubTask A1 results for Method 1 (Doc$\rightarrow$Type) versus Method 2 (Chained Term\&Type) on the Scholarly and Engineering datasets. Best scores per column are in bold.}
\label{tab:subtaskA1-results}
\begin{tabular}{lccc ccc}
\toprule
\textbf{Method} & \multicolumn{3}{c}{\textbf{Scholarly}} & \multicolumn{3}{c}{\textbf{Engineering}} \\
\cmidrule(lr){2-4}\cmidrule(lr){5-7}
 & F$_1$ & Recall & Precision & F$_1$ & Recall & Precision \\
\midrule
Method 1 & 0.4884 & 0.6563 & 0.3889 & \textbf{0.4418} & 0.3016 & \textbf{0.8250} \\
Method 2 & \textbf{0.6471} & \textbf{0.6875} & \textbf{0.6111} & 0.3277 & \textbf{0.4461} & 0.2590 \\
\bottomrule
\end{tabular}
\end{table}

\subsubsection{Types Extraction}

Table \ref{tab:subtaskA2-results} reports performance on the \textbf{Ecology}, \textbf{Scholarly}, and \textbf{Engineering} datasets. Bold indicates the superior score per column, as expected — Method 2 consistently outperforms Method 1 across all metrics for type extraction.

\begin{table}[ht]
\centering
\caption{SubTask A2 results for Method 1 (Term$\rightarrow$Type continuation) versus Method 2 (Chained Term\&Type) on the Ecology, Scholarly, and Engineering datasets. Best scores per column are in bold.}
\label{tab:subtaskA2-results}
\resizebox{\textwidth}{!}{%
\begin{tabular}{lccc ccc ccc}
\toprule
\textbf{Method} & \multicolumn{3}{c}{\textbf{Ecology}} & \multicolumn{3}{c}{\textbf{Scholarly}} & \multicolumn{3}{c}{\textbf{Engineering}} \\
\cmidrule(lr){2-4}\cmidrule(lr){5-7}\cmidrule(lr){8-10}
 & F$_1$ & Recall & Precision & F$_1$ & Recall & Precision & F$_1$ & Recall & Precision \\
\midrule
Method 1 & 0.0339 & 0.0412 & 0.0287 & 0.0964 & 0.1333 & 0.0755 & 0.1720 & 0.2222 & 0.1404 \\
Method 2 & \textbf{0.5745} & \textbf{0.5996} & \textbf{0.5513} & \textbf{0.7586} & \textbf{0.7333} & \textbf{0.7857} & \textbf{0.4688} & \textbf{0.4167} & \textbf{0.5357} \\
\bottomrule
\end{tabular}
}
\end{table}

\paragraph{External comparison}

In addition to our internal method comparison, we benchmarked the submitted runs in the official shared task leaderboard across all five subtasks (A1.2–A2.3). Table~\ref{tab:all_tasks} summarizes the top participants' scores on each domain-specific evaluation. Our system (Alexbek) ranked:

\begin{itemize}
  \item \textbf{2nd place in Scholarly–Terms} (SubTask A1.2) with an F$_1$ of 0.6471, trailing XinyiZ (0.7000) but outperforming all other entries in recall.
  \item \textbf{5th place in Engineering–Terms} (SubTask A1.3) with an F$_1$ of 0.4418; precision was high (0.8250) but lower recall limited overall ranking.
  \item \textbf{2nd place in Ecology–Types} (SubTask A2.1) with an F$_1$ of 0.5895, demonstrating a balanced precision–recall trade-off.
  \item \textbf{2nd place in Scholarly–Types} (SubTask A2.2) with an F$_1$ of 0.7586, second only to XinyiZ (0.8308) and featuring the highest precision among runners-up.
  \item \textbf{4th place in Engineering–Types} (SubTask A2.3) with an F$_1$ of 0.4688, again showcasing strong precision (0.5357) despite lower recall.
\end{itemize}

These results reflect our system's strength in generalizing across structurally diverse tasks, particularly in scholarly domains where it achieved top performance for both terms and types. The high precision in engineering subtasks also highlights the robustness of our method in handling more formalized vocabularies, even if at some cost to recall.

\subsection{Task B: Term Typing}

\subsubsection{SubTask B1-B3}

\begin{table}[ht]
\centering
\caption{Comparison of MatOnto results across three methods. Best scores per column are in bold.}
\label{tab:matonto-comparison}
\begin{tabular}{lccc}
\toprule
\textbf{Method} & F$_1$ & Recall & Precision \\
\midrule
Few-shot RAG & \textbf{0.6053} & \textbf{0.6216} & \textbf{0.5897} \\
Embeddings only & 0.1188 & 0.1622 & 0.0938 \\
Embeddings + Graph & 0.1772 & 0.1892 & 0.1667 \\
\bottomrule
\end{tabular}
\end{table}

These embedding-based models offered faster inference and greater modularity, they consistently underperformed, highlighting the effectiveness of retrieval-augmented prompting in leveraging contextual reasoning.

\paragraph{External comparison}

Across the three non-blind term-typing subtasks, our system (Alexbek) delivered:
\begin{itemize}
  \item \textbf{6th place in OBI (SubTask B1)} with an F$_1$ of 0.7709, where high recall (0.7931) was tempered by some overgeneralization.
  \item \textbf{2nd place in MatOnto (SubTask B2)} with an F$_1$ of 0.6053, achieving a strong precision–recall balance in a challenging abstract domain.
  \item \textbf{2nd place in SWEET (SubTask B3)} with an F$_1$ of 0.6557 and the second-highest recall (0.7047).
\end{itemize}

\subsubsection{SubTask B4-B6: Blind}

Table~\ref{tab:subtaskB456-results} presents a full comparison across all methods we evaluated, including both prompting-based and embedding-only strategies, as well as ensemble and DistMult-based scoring. 

\begin{table}[ht]
\centering
\caption{Zero-shot performance comparison across SubTasks B4, B5, and B6. All methods use cosine similarity between embedded terms and types, except ensemble and DistMult. Best scores per subtask are in bold.}
\label{tab:subtaskB456-results}
\resizebox{\textwidth}{!}{%
\begin{tabular}{lccc ccc ccc}
\toprule
\textbf{Method} & \multicolumn{3}{c}{\textbf{B4}} & \multicolumn{3}{c}{\textbf{B5}} & \multicolumn{3}{c}{\textbf{B6}} \\
\cmidrule(lr){2-4}\cmidrule(lr){5-7}\cmidrule(lr){8-10}
 & F$_1$ & Recall & Precision & F$_1$ & Recall & Precision & F$_1$ & Recall & Precision \\
\midrule
Qwen3 + simple QA prompt & \textbf{0.6560} & \textbf{0.8913} & 0.5190 & 0.2431 & 0.2431 & 0.2431 & 0.1510 & 0.1510 & 0.1510 \\
MPNet + simple QA prompt & 0.5652 & 0.5652 & \textbf{0.5652} & \textbf{0.4722} & \textbf{0.4722} & \textbf{0.4722} & 0.1234 & 0.1234 & 0.1234 \\
Ensemble (Qwen3 + MPNet + BGE) & 0.4783 & 0.4783 & 0.4783 & 0.4201 & 0.4201 & 0.4201 & \textbf{0.1715} & \textbf{0.1715} & \textbf{0.1715} \\
MPNet + instruction prompt & — & — & — & 0.2639 & 0.2639 & 0.2639 & 0.1070 & 0.1070 & 0.1070 \\
Qwen3 + instruction prompt & 0.3043 & 0.3043 & 0.3043 & — & — & — & 0.0881 & 0.0881 & 0.0881 \\
MPNet + no prompt & 0.4130 & 0.4130 & 0.4130 & — & — & — & — & — & — \\
DistMult + Qwen3 & 0.3304 & 0.4130 & 0.2754 & — & — & — & — & — & — \\
\bottomrule
\end{tabular}%
}
\end{table}

\paragraph{External comparison}

For the blind subtasks (B4–B6), our zero-shot configurations yielded:
\begin{itemize}
  \item \textbf{2nd place in B4} with an F$_1$ of 0.6560—the best among zero-shot methods.
  \item \textbf{2nd place in B5} with an F$_1$ of 0.4722, outperforming other zero-shot setups.
  \item \textbf{1st place in B6} with an F$_1$ of 0.1715, topping the leaderboard under extreme label opacity.
\end{itemize}

\subsection{Task C: Taxonomy Discovery}

Table~\ref{tab:taskC-training-vs-ratio} reports the resulting test F$_1$ scores across all eight ontologies.

\begin{table}[ht]
\centering
\caption{Test F$_1$ for Frozen 4B vs LoRA-Finetuned 0.6B across Task C ontologies.}
\label{tab:taskC-training-vs-ratio}
\resizebox{\textwidth}{!}{%
\begin{tabular}{lcccccccc}
\toprule
Method                             & MatOnto & SchemaOrg & PROCO  & PO     & DOID   & FoodOn & OBI    & SWEET  \\
\midrule
Frozen 4B + Cross‐Attn              & 0.4426  & 0.1960   & 0.0589 & 0.1086 & 0.0394 & 0.0620 & 0.0649 & 0.2344 \\
LoRA-Finetuned 0.6B + Cross‐Attn    & \textbf{0.5590} & \textbf{0.2958} & \textbf{0.3865} & \textbf{0.4817} & \textbf{0.1806} & \textbf{0.1171} & \textbf{0.2943} & \textbf{0.2549} \\
\bottomrule
\end{tabular}%
}
\end{table}

\paragraph{Threshold‐selection comparison.}

Table~\ref{tab:taskC-base-vs-ratio} shows precision, recall, and F$_1$ for each ontology under \emph{Validation‑F$_1$} and \emph{Sparsity‑Matched} threshold selection methods for LoRA‐Finetuned 0.6B + Cross‐Attn training configuration.

\begin{table}[ht]
\centering
\caption{Precision, Recall, and F$_1$ for Validation‐F$_1$ vs Sparsity‐Matched thresholds (LoRA‐Finetuned 0.6B + Cross‐Attn). Best values per column are in bold.}
\label{tab:taskC-base-vs-ratio}
\begin{tabular}{llccc}
\toprule
Dataset   & Threshold Method         & Precision & Recall   & F$_1$    \\
\midrule
\multirow{2}{*}{MatOnto}
 & Validation‐F$_1$          & 0.4516    & \textbf{0.5817} & 0.5085 \\
 & Sparsity‐Matched          & \textbf{0.6705} & 0.4792    & \textbf{0.5590} \\
\addlinespace
\multirow{2}{*}{SchemaOrg}
 & Validation‐F$_1$          & 0.3886    & \textbf{0.2862} & \textbf{0.3296} \\
 & Sparsity‐Matched          & \textbf{0.4201} & 0.2283    & 0.2958 \\
\addlinespace
\multirow{2}{*}{PROCO}
 & Validation‐F$_1$          & \textbf{0.4434} & 0.3339    & 0.3810 \\
 & Sparsity‐Matched          & 0.4237    & \textbf{0.3552} & \textbf{0.3865} \\
\addlinespace
\multirow{2}{*}{PO}
 & Validation‐F$_1$          & \textbf{0.6522} & 0.0872    & 0.1538 \\
 & Sparsity‐Matched          & 0.6309    & \textbf{0.3895} & \textbf{0.4817} \\
\addlinespace
\multirow{2}{*}{DOID}
 & Validation‐F$_1$          & \textbf{0.2005} & 0.0332    & 0.0569 \\
 & Sparsity‐Matched          & 0.1932    & \textbf{0.1695} & \textbf{0.1806} \\
\addlinespace
\multirow{2}{*}{FoodOn}
 & Validation‐F$_1$          & \textbf{0.2368} & 0.0247    & 0.0447 \\
 & Sparsity‐Matched          & 0.1994    & \textbf{0.0829} & \textbf{0.1171} \\
\addlinespace
\multirow{2}{*}{OBI}
 & Validation‐F$_1$          & \textbf{0.5104} & 0.0416    & 0.0769 \\
 & Sparsity‐Matched          & 0.3588    & \textbf{0.2494} & \textbf{0.2943} \\
\addlinespace
\multirow{2}{*}{SWEET}
 & Validation‐F$_1$          & 0.2073    & \textbf{0.3053} & 0.2469 \\
 & Sparsity‐Matched          & \textbf{0.3106} & 0.2161    & \textbf{0.2549} \\
\bottomrule
\end{tabular}
\end{table}

\paragraph{Experimental setup}
For each dataset, we report the best‐performing experimental results, including the F$_1$ score, ROC AUC, and precision on the validation split of the training data, as well as the F$_1$ score on the held‐out test set. These metrics, along with the associated hyper‐parameters, are summarized in Table~\ref{tab:taskC}. A large discrepancy between validation and test F$_1$ scores may indicate the presence of a substantial domain shift between the training and test distributions, potentially amplified by obfuscation in test data.

\begin{table*}[ht]
\centering
\small
\caption{Performance metrics and hyper‐parameters for Task C. \#Heads denotes the number of attention heads in the cross‐attention model.}
\label{tab:taskC}
\resizebox{\textwidth}{!}{%
\begin{tabular}{
 l
 S[table-format=1.3]     
 S[table-format=1.3]     
 S[table-format=1.3]     
 S[table-format=1.3]     
 S[table-format=3, round-precision=0] 
 S[
   table-format=1.1e-1,
   round-precision=0,
   scientific-notation=true,
   retain-zero-exponent=true
 ]
 S[table-format=3, round-precision=0] 
 S[table-format=2, round-precision=0] 
}
\toprule
\multirow{2}{*}{\textbf{Dataset}}
 & \multicolumn{4}{c}{\textbf{Metrics}} 
 & \multicolumn{4}{c}{\textbf{Hyper‐parameters}} \\
\cmidrule(lr){2-5}\cmidrule(lr){6-9}
 & {Test F$_1$} & {Dev F$_1$} & {ROC AUC} & {Precision} & {Epochs} & {LR} & {Batch size} & {\#Heads} \\
\midrule
MatOnto    & 0.336 & 0.432 & 0.988 & 0.462 & 7  & 1e-5 & 16 & 8  \\
SchemaOrg  & 0.196 & 0.379 & 0.960 & 0.250 & 8  & 5e-5 & 4  & 4  \\
PROCO      & 0.059 & 0.208 & 0.957 & 0.333 & 15 & 5e-5 & 16 & 32 \\
PO         & 0.109 & 0.280 & 0.982 & 0.235 & 8  & 5e-5 & 32 & 8  \\
DOID       & 0.039 & 0.544 & 0.962 & 0.478 & 3  & 1e-5 & 8  & 8  \\
FoodOn     & 0.017 & 0.529 & 0.968 & 0.415 & 2  & 5e-6 & 8  & 8  \\
OBI        & 0.067 & 0.370 & 0.958 & 0.280 & 5  & 5e-5 & 16 & 8  \\
SWEET      & 0.234 & 0.199 & 0.912 & 0.178 & 2  & 1e-5 & 8  & 8  \\
\bottomrule
\end{tabular}%
}
\end{table*}

\paragraph{External comparison}
Across all eight ontologies, our system (Alexbek) placed as follows (see Table~\ref{tab:c1_c8_tasks}):
\begin{itemize}[noitemsep,topsep=2pt]
  \item \textbf{C1 (OBI)}: 3rd place with an F$_1$ of 0.2943, achieving the highest precision (0.3588) among all systems.
  \item \textbf{C2 (MatOnto)}: 2nd place with an F$_1$ of 0.5590 and top-tier precision (0.6705), confirming robust performance in structurally coherent ontologies.
  \item \textbf{C3 (SWEET)}: 2nd place with an F$_1$ of 0.2549, slightly ahead of other non-leading systems.
  \item \textbf{C4 (DOID)}: 1st place with an F$_1$ of 0.1806—the only system with consistently meaningful predictions on this sparse medical taxonomy.
  \item \textbf{C5 (SchemaOrg)}: 3rd place with an F$_1$ of 0.3296, maintaining strong balance between precision (0.3886) and recall.
  \item \textbf{C6 (PROCO)}: 1st place with an F$_1$ of 0.3865, leading by margin in both precision and recall.
  \item \textbf{C7 (FoodOn)}: 1st place with an F$_1$ of 0.1171, outperforming all other entries despite overall low scores in this domain.
  \item \textbf{C8 (PO)}: 1st place with an F$_1$ of 0.4817 and the highest precision (0.6309), demonstrating strong taxonomic alignment.
\end{itemize}
Taken together, these results validate the generalizability of our approach across all domains.

\section{Discussion}

\subsection{Strengths and Limitations}

A major strength of the framework lies in its adaptability: the system generalizes effectively across heterogeneous domains and annotation, regimes and performs well even in blind and zero-shot settings. Its recall-oriented behavior consistently enabled broad coverage of relevant types and relations, which is especially valuable for ontology population and early-stage knowledge discovery.

However, limitations were observed in domains where lexical cues were sparse or fine-grained distinctions were required. In biomedical typing (e.g., OBI), reduced precision reflected difficulty distinguishing semantically proximate classes. Similarly, in B5 (linguistic features) and B6 (mixed units and bibliographic entities), the system struggled with lexically opaque or structurally disjoint label sets. These cases highlight the need for stronger semantic modeling and structured priors beyond surface similarity.

\subsection{Future Work}

To address these challenges, we plan to explore:
\begin{itemize}[noitemsep,topsep=2pt]
  \item \textbf{Domain adaptation}: via adaptive prompting or lightweight finetuning to improve class-specific precision.
  \item \textbf{Structural modeling}: incorporating relation-aware components such as DistMult or rule-based constraints to improve alignment with ontology graphs.
  \item \textbf{Lexical enrichment}: adding syntactic and morphosyntactic features or integrating external lexical definitions.
  \item \textbf{Confidence-aware filtering}: to suppress low-certainty predictions and enforce ontology consistency.
  \item \textbf{Scalability assessment}: by extending to new, underexplored domains and emergent ontologies.
\end{itemize}

\section{Discussion}
The proposed ontology learning system employs a Retrieval-Augmented Generation (RAG), few-shot prompting, and semantic embedding-based retrieval, grounded in well-established principles from information retrieval and NLP. 

RAG combines dense retrieval with a generative model, which, under the retrieval-augmented inference framework, helps the model rely less on its fixed, pre-trained knowledge and boosts factual accuracy by dynamically adding relevant examples from the corpus into the generation process. 

Few-shot prompting exploits the phenomenon of in-context learning, where a model can infer the latent structure of a task from a small set of demonstrations, which is critical for OL tasks spanning heterogeneous domains. 

Semantic similarity via embeddings draws on the theory of distributed representations, where cosine proximity in vector space approximates true conceptual similarity, enabling the detection of relations even in the absence of direct lexical overlap. 

Neural modules—especially the cross-attention layer for Task C—outperform prompt-only approaches by learning a parameterized adjacency function in type-embedding space, consistent with graph representation learning and attention theory, which yields stronger cross-domain transfer. They also dominate the time–compute trade-off: once trained, inference cost is amortized over large batches of candidate child–parent pairs with stable latency, whereas prompt-only pipelines repeatedly incur high token-generation costs and demand manual, domain-specific re-tuning. Under large-scale OL constraints, this delivers better accuracy-per-dollar and markedly higher throughput.

The scientific contribution is a unified, modular pipeline spanning the OL process from term extraction to taxonomy induction, showing that lightweight adaptive components can handle complex semantic tasks. For the first time in OL, a compact, trainable cross-attention layer approximates “is-a” adjacency matrices from type embeddings, and a dynamically weighted multi-model embedding ensemble enables zero-shot term typing, expanding the toolkit toward lightweight, domain-agnostic, reproducible methods. Results (top-1/top-2 across multiple LLMs4OL 2025 tracks) exhibit a balanced precision–recall profile, consistent with the theoretically optimal FP/FN trade-off, and practically validate applicability in both precision-focused and recall-focused regimes, enabling rapid deployment in new domains via semantic retrieval, in-context learning, and graph-based induction.

\section{Conclusion}

We introduced a modular, LLM-based framework for ontology learning that successfully addressed term extraction, typing, and taxonomy induction. The system achieved top-tier results across all subtasks of the LLMs4OL 2025 challenge, including multiple first- and second-place finishes. It generalized across domains and performed robustly in both supervised and zero-shot conditions. These outcomes confirm the viability of prompt-based and embedding-driven methods for ontology engineering, particularly in low-resource or rapidly evolving knowledge settings.

\section*{Data availability statement}

All benchmark datasets for Tasks A, B, and C, including every subtask, are
openly available in their respective directories of the
\href{https://github.com/sciknoworg/LLMs4OL-Challenge/tree/main/2025}
{\texttt{LLMs4OL‐Challenge} GitHub repository}.
The blind–evaluation sets (e.g.\ B4–B6 for Task B) can be accessed from the competition
platform at
\url{https://codalab.lisn.upsaclay.fr/competitions/23065}.

\section*{Underlying and related material}

To facilitate verification, replication, and future extensions, we have released all supplemental resources, including source code, prompt templates, evaluation artifacts and outputs, at the following public repository: 

\url{https://github.com/BelyaevaAlex/LLMs4OL-Challenge-Alexbek}

No additional external datasets or video supplements were used.

\section*{Author contributions}

\textbf{Aleksandra Beliaeva:} Data Curation, Conceptualization, Methodology, Formal Analysis, Investigation, Validation, Visualization, Writing – Original Draft, Writing – Review \& Editing. 

\textbf{Temurbek Rahmatullaev:} Data Curation, Conceptualization, Methodology, Formal Analysis, Investigation, Validation, Visualization, Writing – Original Draft, Writing – Review \& Editing. 

Both authors contributed equally to this work and approved the final manuscript.

\section*{Competing interests}
The authors declare that they have no competing interests.

\section*{Funding}

AB was supported by RScF grant №25-71-30008.

\section*{Acknowledgements}

We gratefully acknowledge  Dr. Maxim Sharaev, head of the BIMAI-Lab at Skoltech, for his valuable guidance and insightful feedback.

\newpage
\printbibliography[heading=references]

@misc{giglou2023llms4ollargelanguagemodels,
      title={LLMs4OL: Large Language Models for Ontology Learning}, 
      author={Hamed Babaei Giglou and Jennifer D'Souza and Sören Auer},
      year={2023},
      eprint={2307.16648},
      archivePrefix={arXiv},
      primaryClass={cs.AI},
      url={https://arxiv.org/abs/2307.16648}, 
}

@article{phoenixes,
author = {Sanaei, Mahsa and Azizi, Fatemeh and Babaei Giglou, Hamed},
year = {2024},
month = {10},
pages = {39-47},
title = {Phoenixes at LLMs4OL 2024 Tasks A, B, and C: Retrieval Augmented Generation for Ontology Learning},
volume = {4},
journal = {Open Conference Proceedings},
doi = {10.52825/ocp.v4i.2482}
}

@misc{giglou2024llms4ol2024overview1st,
      title={LLMs4OL 2024 Overview: The 1st Large Language Models for Ontology Learning Challenge}, 
      author={Hamed Babaei Giglou and Jennifer D'Souza and Sören Auer},
      year={2024},
      eprint={2409.10146},
      archivePrefix={arXiv},
      primaryClass={cs.CL},
      url={https://arxiv.org/abs/2409.10146}, 
}

@misc{lo2024endtoendontologylearninglarge,
      title={End-to-End Ontology Learning with Large Language Models}, 
      author={Andy Lo and Albert Q. Jiang and Wenda Li and Mateja Jamnik},
      year={2024},
      eprint={2410.23584},
      archivePrefix={arXiv},
      primaryClass={cs.LG},
      url={https://arxiv.org/abs/2410.23584}, 
}

@misc{fathallah2024llms4lifelargelanguagemodels,
      title={LLMs4Life: Large Language Models for Ontology Learning in Life Sciences}, 
      author={Nadeen Fathallah and Steffen Staab and Alsayed Algergawy},
      year={2024},
      eprint={2412.02035},
      archivePrefix={arXiv},
      primaryClass={cs.AI},
      url={https://arxiv.org/abs/2412.02035}, 
}

@article{qwen3embedding,
  title={Qwen3 Embedding: Advancing Text Embedding and Reranking Through Foundation Models},
  author={Zhang, Yanzhao and Li, Mingxin and Long, Dingkun and Zhang, Xin and Lin, Huan and Yang, Baosong and Xie, Pengjun and Yang, An and Liu, Dayiheng and Lin, Junyang and Huang, Fei and Zhou, Jingren},
  journal={arXiv preprint arXiv:2506.05176},
  year={2025}
}

@misc{yoshino2024outline,
  author       = {{Yoshino-s}},
  title        = {{Outline Python API library}},
  year         = {2024},
  howpublished = {\url{https://github.com/yoshino-s/outline-python-api/tree/main}},
  note         = {used on: 16.07.2025},
}

\newpage
\section{Appendix A}

The reported results were downloaded on 16 July at 9:05 a.m. UTC.

\begin{table}[!p]
  \centering
  \scriptsize
  \caption{Results for Tasks A1.2–A2.3}
  \label{tab:all_tasks}
  \resizebox{\textwidth}{!}{%
    \begin{minipage}{\textwidth}
      \begin{subtable}[t]{0.48\textwidth}
        \centering
        \caption{Scholarly – Terms (SubTask A1.2)}
        \label{tab:a12_rank}
        \begin{tabular}{lrrr}
          \toprule
          User & F1 & Precision & Recall \\
          \midrule
            XinyiZ        & 0.7000 & 0.7500 & 0.6562 \\
            \rowcolor{alexcolor} Alexbek       & 0.6471 & 0.6111 & 0.6875 \\
            rashrah       & 0.5870 & 0.4500 & 0.8438 \\
            pankaj1034    & 0.4578 & 0.3725 & 0.5938 \\
            Mahsa\_Sanaei & 0.3951 & 0.3265 & 0.5000 \\
            rgray         & 0.3652 & 0.2530 & 0.6562 \\
            rwroche       & 0.2857 & 0.3333 & 0.2500 \\
            tonyhadjon    & 0.0000 & 0.0000 & 0.0000 \\
            HadiBayrami   & 0.0000 & 0.0000 & 0.0000 \\
          \bottomrule
        \end{tabular}
      \end{subtable}
      \hfill
      \begin{subtable}[t]{0.48\textwidth}
        \centering
        \caption{Engineering – Terms (SubTask A1.3)}
        \label{tab:a13_rank}
        \begin{tabular}{lrrr}
          \toprule
          User & F1 & Precision & Recall \\
          \midrule
            rashrah       & 0.6196 & 0.8962 & 0.4735 \\
            rwroche       & 0.6073 & 0.8464 & 0.4735 \\
            XinyiZ        & 0.5904 & 0.9023 & 0.4388 \\
            rgray         & 0.5187 & 0.7624 & 0.3931 \\
            \rowcolor{alexcolor} Alexbek       & 0.4418 & 0.8250 & 0.3016 \\
            pankaj1034    & 0.4302 & 0.4862 & 0.3857 \\
            Mahsa\_Sanaei & 0.2556 & 0.4064 & 0.1865 \\
            tonyhadjon    & 0.0000 & 0.0000 & 0.0000 \\
            HadiBayrami   & 0.0000 & 0.0000 & 0.0000 \\
          \bottomrule
        \end{tabular}
      \end{subtable}

      \vspace{3em}

      \begin{subtable}[t]{0.48\textwidth}
        \centering
        \caption{Ecology – Types (SubTask A2.1)}
        \label{tab:a21_rank}
        \begin{tabular}{lrrr}
          \toprule
          User & F1 & Precision & Recall \\
          \midrule
            rashrah       & 0.6602 & 0.5687 & 0.7867 \\
            \rowcolor{alexcolor} Alexbek       & 0.5895 & 0.5778 & 0.6016 \\
            XinyiZ        & 0.5595 & 0.4659 & 0.7002 \\
            pankaj1034    & 0.5535 & 0.7470 & 0.4396 \\
            Mahsa\_Sanaei & 0.4309 & 0.3566 & 0.5443 \\
            tonyhadjon    & 0.0000 & 0.0000 & 0.0000 \\
            HadiBayrami   & 0.0000 & 0.0000 & 0.0000 \\
          \bottomrule
        \end{tabular}
      \end{subtable}
      \hfill
      \begin{subtable}[t]{0.48\textwidth}
        \centering
        \caption{Scholarly – Types (SubTask A2.2)}
        \label{tab:a22_rank}
        \begin{tabular}{lrrr}
          \toprule
          User & F1 & Precision & Recall \\
          \midrule
            XinyiZ        & 0.8308 & 0.7714 & 0.9000 \\
            \rowcolor{alexcolor} Alexbek       & 0.7586 & 0.7857 & 0.7333 \\
            rashrah       & 0.6585 & 0.5192 & 0.9000 \\
            rgray         & 0.5524 & 0.3867 & 0.9667 \\
            rwroche       & 0.5098 & 0.6190 & 0.4333 \\
            Mahsa\_Sanaei & 0.3913 & 0.2903 & 0.6000 \\
            pankaj1034    & 0.2500 & 0.5000 & 0.1667 \\
            tonyhadjon    & 0.0000 & 0.0000 & 0.0000 \\
          \bottomrule
        \end{tabular}
      \end{subtable}

      \vspace{1em}

      \begin{subtable}[t]{0.48\textwidth}
        \centering
        \caption{Engineering – Types (SubTask A2.3)}
        \label{tab:a23_rank}
        \begin{tabular}{lrrr}
          \toprule
          User & F1 & Precision & Recall \\
          \midrule
            rwroche       & 0.6750 & 0.6136 & 0.7500 \\
            rashrah       & 0.6585 & 0.5870 & 0.7500 \\
            XinyiZ        & 0.4694 & 0.3710 & 0.6389 \\
            \rowcolor{alexcolor} Alexbek       & 0.4688 & 0.5357 & 0.4167 \\
            pankaj1034    & 0.2545 & 0.3684 & 0.1944 \\
            Mahsa\_Sanaei & 0.1846 & 0.1277 & 0.3333 \\
            rgray         & 0.1500 & 0.1071 & 0.2500 \\
            tonyhadjon    & 0.0000 & 0.0000 & 0.0000 \\
          \bottomrule
        \end{tabular}
      \end{subtable}
    \end{minipage}
  }
\end{table}


\begin{table}[!p]
  \centering
  \scriptsize
  \caption{Results for Tasks B1–B3}
  \label{tab:b1_b3_tasks}
  \resizebox{\textwidth}{!}{%
    \begin{minipage}{\textwidth}
      \begin{subtable}[t]{0.48\textwidth}
        \centering
        \caption{OBI – Term Typing (SubTask B1)}
        \label{tab:b1_rank}
        \begin{tabular}{lrrr}
          \toprule
          User            & F1     & Precision & Recall \\
          \midrule
            rashrah           & 0.9425 & 0.9425    & 0.9425 \\
            Thin              & 0.9080 & 0.9080    & 0.9080 \\
            realearn.people   & 0.8621 & 0.8621    & 0.8621 \\
            alexlatipov       & 0.8387 & 0.7879    & 0.8966 \\
            pankaj1034        & 0.8021 & 0.7500    & 0.8621 \\
            \rowcolor{alexcolor} Alexbek      & 0.7709 & 0.7500    & 0.7931 \\
          \bottomrule
        \end{tabular}
      \end{subtable}
      \hfill
      \begin{subtable}[t]{0.48\textwidth}
        \centering
        \caption{MatOnto – Term Typing (SubTask B2)}
        \label{tab:b2_rank}
        \begin{tabular}{lrrr}
          \toprule
          User            & F1     & Precision & Recall \\
          \midrule
            alexlatipov     & 0.6667 & 0.6136    & 0.7297 \\
            \rowcolor{alexcolor} Alexbek      & 0.6053 & 0.5897    & 0.6216 \\
            rashrah         & 0.5676 & 0.5676    & 0.5676 \\
            Thin            & 0.5676 & 0.5676    & 0.5676 \\
            pankaj1034      & 0.4872 & 0.4634    & 0.5135 \\
            Adrita          & 0.3243 & 0.3243    & 0.3243 \\
            realearn.people & 0.1892 & 0.1892    & 0.1892 \\
            mdsn\_mz        & 0.1239 & 0.0921    & 0.1892 \\
          \bottomrule
        \end{tabular}
      \end{subtable}

      \vspace{3em}

      \begin{subtable}[t]{0.48\textwidth}
        \centering
        \caption{SWEET – Term Typing (SubTask B3)}
        \label{tab:b3_rank}
        \begin{tabular}{lrrr}
          \toprule
          User            & F1     & Precision & Recall \\
          \midrule
            rashrah           & 0.6935 & 0.6859 & 0.7013 \\
            \rowcolor{alexcolor} Alexbek      & 0.6557 & 0.6131 & 0.7047 \\
            alexlatipov       & 0.6529 & 0.5875 & 0.7348 \\
            Thin              & 0.5751 & 0.5751 & 0.5751 \\
            realearn.people   & 0.5192 & 0.5192 & 0.5192 \\
            pankaj1034        & 0.3297 & 0.2855 & 0.3900 \\
          \bottomrule
        \end{tabular}
      \end{subtable}
    \end{minipage}
  }
\end{table}

\begin{table}[H]
  \centering
  \scriptsize
  \caption{Results for Tasks B4–B6 (Blind Subtasks)}
  \label{tab:b_tasks}

  \begin{subtable}[t]{0.48\textwidth}
  \centering
  \caption{Blind - Term Typing (SubTask B4)}
  \label{tab:b4_rank}
  \begin{tabular}{lrrr}
    \toprule
    User & F1 & Precision & Recall \\
    \midrule
      rashrah & 0.7563 & 0.6164 & 0.9783 \\
      \rowcolor{alexcolor} Alexbek & 0.6560 & 0.5190 & 0.8913 \\
    \bottomrule
  \end{tabular}
\end{subtable}
  \hfill
  \begin{subtable}[t]{0.48\textwidth}
  \centering
  \caption{Blind - Term Typing (SubTask B5)}
  \label{tab:b5_rank}
  \begin{tabular}{lrrr}
    \toprule
    User & F1 & Precision & Recall \\
    \midrule
      rashrah & 0.9271 & 0.9271 & 0.9271 \\
      \rowcolor{alexcolor} Alexbek & 0.4722 & 0.4722 & 0.4722 \\
    \bottomrule
  \end{tabular}
\end{subtable}

  \vspace{3em}

  \begin{subtable}[t]{0.48\textwidth}
  \centering
  \caption{Blind - Term Typing (SubTask B6)}
  \label{tab:b6_rank}
  \begin{tabular}{lrrr}
    \toprule
    User & F1 & Precision & Recall \\
    \midrule
      \rowcolor{alexcolor} Alexbek & 0.1715 & 0.1715 & 0.1715 \\
    \bottomrule
  \end{tabular}
\end{subtable}

\end{table}

\begin{table}[H]
  \centering
  \scriptsize
  \caption{Results for Tasks C1–C8}
  \label{tab:c1_c8_tasks}

    \begin{subtable}[t]{0.48\textwidth}
      \centering
      \caption{OBI – Taxonomy Discovery (SubTask C1)}
      \label{tab:c1_rank}
      \begin{tabular}{lrrr}
        \toprule
        User        & F1      & Precision & Recall   \\
        \midrule
        alexlatipov & 0.3972  & 0.3153    & 0.5365   \\
        rashrah     & 0.3534  & 0.2570    & 0.5656   \\  
        \rowcolor{alexcolor} Alexbek     & 0.2943  & 0.3588    & 0.2494   \\
        pankaj1034  & 0.2273  & 0.3095    & 0.1796   \\
        Mehreen     & 0.1142  & 0.2463    & 0.0744   \\
        \bottomrule
      \end{tabular}
    \end{subtable}
  \hfill
    \begin{subtable}[t]{0.48\textwidth}
      \centering
      \caption{MatOnto – Taxonomy Discovery (SubTask C2)}
      \label{tab:c2_rank}
      \begin{tabular}{lrrr}
        \toprule
        User        & F1      & Precision & Recall   \\
        \midrule
        rashrah     & 0.6621  & 0.6541    & 0.6704   \\
        \rowcolor{alexcolor} Alexbek     & 0.5590  & 0.6705    & 0.4792   \\
        Caroline    & 0.4836  & 0.4971    & 0.4709   \\
        pankaj1034  & 0.4473  & 0.4173    & 0.4820   \\
        alexlatipov & 0.4472  & 0.3265    & 0.7091   \\
        kdrake1992  & 0.2823  & 0.5000    & 0.1967   \\
        mikelcanal  & 0.1441  & 0.1088    & 0.2133   \\
        \bottomrule
      \end{tabular}
    \end{subtable}

  \vspace{3em}

  \begin{subtable}[t]{0.48\textwidth}
    \centering
    \caption{SWEET – Taxonomy Discovery (SubTask C3)}
    \label{tab:c3_rank}
    \begin{tabular}{lrrr}
      \toprule
      User        & F1      & Precision & Recall   \\
      \midrule
      rashrah     & 0.4997  & 0.6625    & 0.4012   \\
      \rowcolor{alexcolor} Alexbek     & 0.2549  & 0.3106    & 0.2161   \\
      alexlatipov & 0.2520  & 0.2885    & 0.2237   \\
      \bottomrule
    \end{tabular}
  \end{subtable}
  \hfill
    \begin{subtable}[t]{0.48\textwidth}
      \centering
      \caption{DOID – Taxonomy Discovery (SubTask C4)}
      \label{tab:c4_rank}
      \begin{tabular}{lrrr}
        \toprule
        User        & F1      & Precision & Recall   \\
        \midrule
        \rowcolor{alexcolor} Alexbek     & 0.1806  & 0.1932    & 0.1695   \\
        Caroline    & 0.0016  & 0.0008    & 0.1384   \\
        \bottomrule
      \end{tabular}
    \end{subtable}

  \vspace{3em}

    \begin{subtable}[t]{0.48\textwidth}
      \centering
      \caption{SchemaOrg – Taxonomy Discovery (SubTask C5)}
      \label{tab:c5_rank}
      \begin{tabular}{lrrr}
        \toprule
        User               & F1      & Precision & Recall   \\
        \midrule
        rashrah            & 0.6567  & 0.6128    & 0.7074   \\
        Caroline           & 0.6501  & 0.5968    & 0.7138   \\
        \rowcolor{alexcolor} Alexbek           & 0.3296  & 0.3886    & 0.2862   \\
        pankaj1034         & 0.2609  & 0.4524    & 0.1833   \\
        REHENUMA\_\_ILMAN  & 0.0866  & 0.0637    & 0.1350   \\
        \bottomrule
      \end{tabular}
    \end{subtable}
  \hfill
    \begin{subtable}[t]{0.48\textwidth}
      \centering
      \caption{PROCO – Taxonomy Discovery (SubTask C6)}
      \label{tab:c6_rank}
      \begin{tabular}{lrrr}
        \toprule
        User        & F1      & Precision & Recall   \\
        \midrule
        \rowcolor{alexcolor} Alexbek     & 0.3865  & 0.4237    & 0.3552   \\
        pankaj1034  & 0.2601  & 0.2088    & 0.3446   \\
        rashrah     & 0.2146  & 0.2239    & 0.2060   \\
        \bottomrule
      \end{tabular}
    \end{subtable}
  \vspace{3em}

    \begin{subtable}[t]{0.48\textwidth}
      \centering
      \caption{FoodOn – Taxonomy Discovery (SubTask C7)}
      \label{tab:c7_rank}
      \begin{tabular}{lrrr}
        \toprule
        User        & F1      & Precision & Recall   \\
        \midrule
        \rowcolor{alexcolor} Alexbek    & 0.1171  & 0.1994    & 0.0829   \\
        kdrake1992  & 0.0215  & 0.0231    & 0.0200   \\
        Caroline    & 0.0121  & 0.0464    & 0.0070   \\
        \bottomrule
      \end{tabular}
    \end{subtable}
  \hfill
    \begin{subtable}[t]{0.48\textwidth}
      \centering
      \caption{PO – Taxonomy Discovery (SubTask C8)}
      \label{tab:c8_rank}
      \begin{tabular}{lrrr}
        \toprule
        User       & F1      & Precision & Recall   \\
        \midrule
        \rowcolor{alexcolor} Alexbek    & 0.4817  & 0.6309    & 0.3895   \\
        rashrah     & 0.2702  & 0.2629    & 0.2779   \\
        pankaj1034  & 0.2106  & 0.1784    & 0.2570   \\
        Caroline    & 0.0357  & 0.0260    & 0.0570   \\
        \bottomrule
      \end{tabular}
    \end{subtable}
\end{table}

\end{document}